\documentclass[sigconf,9pt]{acmart}
\usepackage{multirow}
\usepackage{makecell}
\usepackage{balance}
\usepackage{ulem}
\usepackage{url}
\newtheorem{definition}{Definition}[section]
\AtBeginDocument{%
  }

\copyrightyear{2025} 
\acmYear{2025} 
\setcopyright{acmlicensed}\acmConference[SIGMOD-Companion
'25]{Companion of the 2025 International Conference on Management of
Data}{June 22--27, 2025}{Berlin, Germany}
\acmBooktitle{Companion of the 2025 International Conference on
Management of Data (SIGMOD-Companion '25), June 22--27, 2025, Berlin, Germany}
\acmDOI{10.1145/3626246.3654733}
\acmISBN{979-8-4007-0422-2/24/06}

\begin{document}

\title[KDSelector: A Knowledge-Enhanced and Data-Efficient Model Selector Learning Framework for TSAD]{KDSelector: A Knowledge-Enhanced and Data-Efficient Model Selector Learning Framework for Time Series Anomaly Detection}

\author{Zhiyu Liang}
\orcid{0000-0003-0083-2547}
\affiliation{%
	\institution{Harbin Institute of Technology}
	\city{Harbin}
	\country{China}
}
\email{zyliang@hit.edu.cn}

\author{Dongrui Cai}
\authornote{Equal contribution.}
\email{2021113243@stu.hit.edu.cn}
\orcid{0009-0005-4036-0806}
\author{Chenyuan Zhang}
\authornotemark[1]
\email{2022113454@stu.hit.edu.cn}
\orcid{0009-0002-5376-391X}
\affiliation{%
  \institution{Harbin Institute of Technology}
  \city{Harbin}
  \country{China}
}

\author{Zheng Liang}
\orcid{0000-0003-1844-4366}
\email{lz20@hit.edu.cn}
\author{Chen Liang}
\email{23B903050@stu.hit.edu.cn}
\orcid{0000-0002-1093-0362}
\affiliation{%
	\institution{Harbin Institute of Technology}
	\city{Harbin}
	\country{China}
}

\author{Bo Zheng}
\orcid{0009-0005-5309-3364}
\affiliation{%
	\institution{CnosDB Inc.}
	\city{Beijing}
	\country{China}
}
\email{harbour.zheng@cnosdb.com}

\author{Shi Qiu}
%\email{sheldon.qiu@csu.edu.cn}
\orcid{0000-0002-8027-3396}
\author{Jin Wang}
\orcid{0000-0003-4889-2717}
\email{{sheldon.qiu, jerryW}@csu.edu.cn}
\affiliation{%
	\institution{Central South University}
	\city{Changsha}
	\country{China}
}

\author{Hongzhi Wang}
\orcid{0000-0002-7521-2871}
\affiliation{%
	\institution{Harbin Institute of Technology}
	\city{Harbin}
	\country{China}
}
\email{wangzh@hit.edu.cn}\authornote{Corresponding author.}

\renewcommand{\shortauthors}{Liang et al.}

\begin{abstract}
Model selection has been raised as an essential problem in the area of time series anomaly detection (TSAD), because there is no single best TSAD model for the highly heterogeneous time series in real-world applications. However, despite the success of existing model selection solutions that train a classification model (especially neural network, NN) using historical data as a selector to predict the correct TSAD model for each series, the NN-based selector learning methods used by existing solutions do not make full use of the knowledge in the historical data and require iterating over all training samples, which limits the accuracy and training speed of the selector. To address these limitations, we propose KDSelector, a novel knowledge-enhanced and data-efficient framework for learning the NN-based TSAD model selector, of which three key components are specifically designed to integrate available knowledge into the selector and dynamically prune less important and redundant samples during the learning. We develop a TSAD model selection system with KDSelector as the internal, to demonstrate how users improve the accuracy and training speed of their selectors by using KDSelector as a plug-and-play module. Our demonstration video is hosted at \url{https://youtu.be/2uqupDWvTF0}.      
\end{abstract}

\begin{CCSXML}
<ccs2012>
   <concept>
       <concept_id>10002950.10003648.10003688.10003693</concept_id>
       <concept_desc>Mathematics of computing~Time series analysis</concept_desc>
       <concept_significance>500</concept_significance>
       </concept>
   <concept>
       <concept_id>10010147.10010257</concept_id>
       <concept_desc>Computing methodologies~Machine learning</concept_desc>
       <concept_significance>500</concept_significance>
       </concept>
 </ccs2012>
\end{CCSXML}

\ccsdesc[500]{Mathematics of computing~Time series analysis}
\ccsdesc[500]{Computing methodologies~Machine learning}

\keywords{Model selection, Time series, Anomaly detection}


\maketitle

\vspace{-1.4ex}
\section{Introduction}\label{sec:intro}
Time series anomaly detection (TSAD) is an important technique for many real-world applications~\cite{schmidl2022anomaly}. However, as widely shown, there is \textbf{\textit{no single best TSAD method (a.k.a. model) when applied to different time series}}, 
 due to the highly heterogeneous nature of the data in terms of the types, numbers, and lasting time of the anomalies, etc~\cite{schmidl2022anomaly,ChooseWisely}.  A straightforward thought is to combine all TSAD models through ensembling. Nevertheless, such solutions require running multiple TSAD methods, causing excessive computational costs that are prohibitive for large time series collections.

To overcome the above issues, recent work~\cite{ChooseWisely} proposes \textbf{\textit{model selection methods to automatically select the best TSAD model for different time series based on their data characteristics}}. This is usually achieved by training (a.k.a. learning) a time series classification~\cite{liang2024units} (TSC) model as a \textbf{selector} to classify time series into discrete categories that represent the TSAD models to select, using the historical data such as the previously seen time series and the corresponding correct TSAD models as training samples. By such solutions, only the selected TSAD model is run for each time series to detect, which is more scalable than the aforementioned ensemble methods that require running all candidates. 

\noindent
\textbf{\underline{Challenges.}} Among existing approaches, neural network (NN)-based selectors (i.e., TSC models) have shown superior accuracy~\cite{ChooseWisely} due to their ability to learn complex relationships between time series and TSAD models. However, the selector learning methods used by the existing solutions face two main challenges.

\begin{figure}[htbp]
    \centering
    \includegraphics[width=0.75\linewidth]{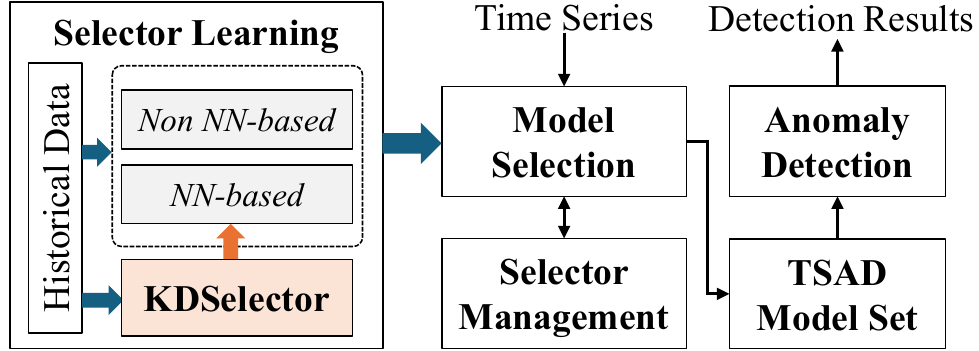}
    \vspace{-1.3ex}
    \caption{System architecture.}
    \label{fig:arch}
    \vspace{-4ex}
\end{figure}

\textit{Firstly}, existing methods only use the time series and the hard labels that represent the best TSAD models for the corresponding series as training data, \textbf{\textit{ignoring that there usually exists additional knowledge in the historical data}}, such as the detection performance of all TSAD model candidates used for identifying the hard labels~\cite{ChooseWisely}, and the metadata that reflects the characteristics of the time series, anomalies, TSAD scenarios, etc~\cite{schmidl2022anomaly}. Thus, the learned selector can be weak in choosing a good TSAD model.    
%However, existing TSAD model selection methods still suffer from insufficient performance in terms of selecting accurate TSAD models and high computational overhead for learning the selector. The main reasons  

\textit{Secondly}, NN-based selector learning in existing solutions, following the widely used stochastic gradient descent (SGD) scheme~\cite{lecun2015deep-learning}, \textbf{\textit{requires iterating over all training samples, which is data-inefficient and time-consuming.}} Although there exists advanced acceleration method~\cite{qin2024infobatch} that can dynamically prune less important training samples for each training epoch while guaranteeing little loss of accuracy, the existing approach has not considered the specific data characteristics in the TSAD model selection problem, which causes it to provide only limited speedup.      

\noindent
\textbf{\underline{Contributions.}} To address the first challenge, we propose \textbf{\textit{two knowledge enhancement modules}} to integrate the knowledge into the selector to improve its selection ability. To utilize the detection performance of all TSAD models, we design a \textbf{\textit{\underline{p}erformance-\underline{i}nformed \underline{s}elector \underline{l}earning (PISL)}} module that transforms the performance scores of different TSAD models into the probabilities of selecting the corresponding models, which is taken as a soft label to better train the selector. To gain knowledge from diverse metadata, we propose a \textbf{\textit{\underline{m}eta-\underline{k}nowledge \underline{i}ntegration (MKI)}} module. The module takes natural language as input, so that any type of metadata can be flexibly incorporated. It transforms the input text into unified embedding (i.e., feature vector) via a pre-trained large language model (LLM)~\cite{bert}. Then, it integrates the knowledge within the input into the selector by maximizing the mutual information between the feature vectors of the text and the time series. 

To cope with the second challenge, we propose \textbf{\textit{a novel \underline{p}runing-based \underline{a}cceleration (PA) framework}} for NN-based selector training that can prune more training samples at each epoch with still nearly lossless model accuracy. Our key observation is that there are training samples that are very similar to each other and also contribute to almost equal training losses. Based on our theoretical analysis, these training samples have redundant information for selector learning. Therefore, we randomly prune them and rescale the gradients of the remaining samples, which not only improves the training speed, but also ensures that training on the pruned dataset can achieve a similar result as training on the original one.     

%To address these challenges, we propose TS-KDS, to the best of our knowledge, the first selector learning framework for TSAD model selection that improves the accuracy and training speed of the NN-based selector via knowledge enhancement and data-efficient  

To the best of our knowledge, our novel solution, which we name \textbf{\textit{\texttt{KDSelector}, is the first framework for NN-based TSAD model \underline{selector} learning that aims to improve the accuracy and training speed via \underline{k}nowledge enhancement and \underline{d}ata pruning}}. It is noteworthy that the three proposed key components, including PISL, MKI, and PA, are all \textbf{\textit{plug-and-play frameworks}} that are \textbf{\textit{agnostic to NN architectures}} (e.g., ResNet or Transformer~\cite{ChooseWisely}) and \textbf{\textit{independent of each other}}. This means that users can flexibly integrate each of them into their own TSAD model selection tasks where any selector architecture can be used.

This paper aims to \textbf{\textit{demonstrate our KDSelector in two aspects}}, including (i) guiding audiences to learn a selector and apply it to TSAD model selection on their own data, and (ii) showcasing the effectiveness and superiority of the proposed methods in improving the accuracy and training speed of the selectors.  To achieve our goal, we develop \textbf{\textit{an end-to-end system}} to enable TSAD model selection using different TSC methods (i.e., selectors), where our \textbf{\textit{KDSelector can be flexibly used for training any NN-based selector}}. Currently, we have implemented \textbf{\textit{12 TSAD models}} and \textbf{\textit{15 selectors}}, and provided \textbf{\textit{16 different datasets}} to facilitate evaluation. Our code is available at \textbf{\textit{\url{https://github.com/chenyuanTKCY/KDSelector}}}.

\vspace{-0.5ex}
\section{System Overview}

\noindent
\textbf{\underline{Preliminaries.}} Formally, the problem of TSAD model selection is defined as follows.

\begin{definition}[TSAD model selection]
Given a set of TSAD models, denoted as $\mathcal{M} = \{M_1,\ldots,M_m\}$, TSAD model selection aims to build a function (i.e., selector) $f$ to predict the model in $\mathcal{M}$ that has the best detection performance for an input time series $T \in \mathbb{R}^L$, i.e.,
\begin{equation}
    f(T) = \underset{i=1,\ldots,m}{\arg\max}\ P(M_i(T)),
\end{equation}
where $P$ can be any interested metric, such as AUC-PR or F1-Score.\label{def:TSADMS}
\end{definition}

Definition~\ref{def:TSADMS} can be seen as a specific TSC problem~\cite{liang2024units}, where the selector $f$ is a TSC model that maps a time series $T$ to a class that represents the TSAD model to select from $\mathcal{M}$. Therefore, we can adopt any existing TSC method to build $f$, using available historical data such as the previously seen time series and the corresponding correct TSAD models as training samples~\cite{ChooseWisely}. To address real-world time series of variable lengths, we follow~\cite{ChooseWisely} to preprocess each raw time series by \textit{extracting fixed-length subsequences} using a window of size $L$. The selector $f$ predicts a TSAD model for each subsequence, while majority voting is used to \textit{select one model for each time series from the predicted models of all its subsequences.}

\noindent
\textbf{\underline{System architecture.}} Fig.~\ref{fig:arch} shows the architecture of our TSAD model selection system, which includes five main components. 

The \textbf{Selector Learning} module, which is the key to model selection, aims to train TSC models as selectors using historical data. Currently, the system supports 15 different selectors, including both NN-based and non-NN-based, where \textbf{\textit{our novel KDSelector, as illustrated in Sect.~\ref{sec:internal}, is used as a plug-and-play framework to improve the learning of any NN-based selector}}. For evaluation and demonstration purposes, we have prepared the 16 different TSAD datasets used in~\cite{ChooseWisely} as historical data. Audiences can also test on their own data using our system.   

\begin{figure*}[htbp]
        \centering
    \includegraphics[width=.9\linewidth]{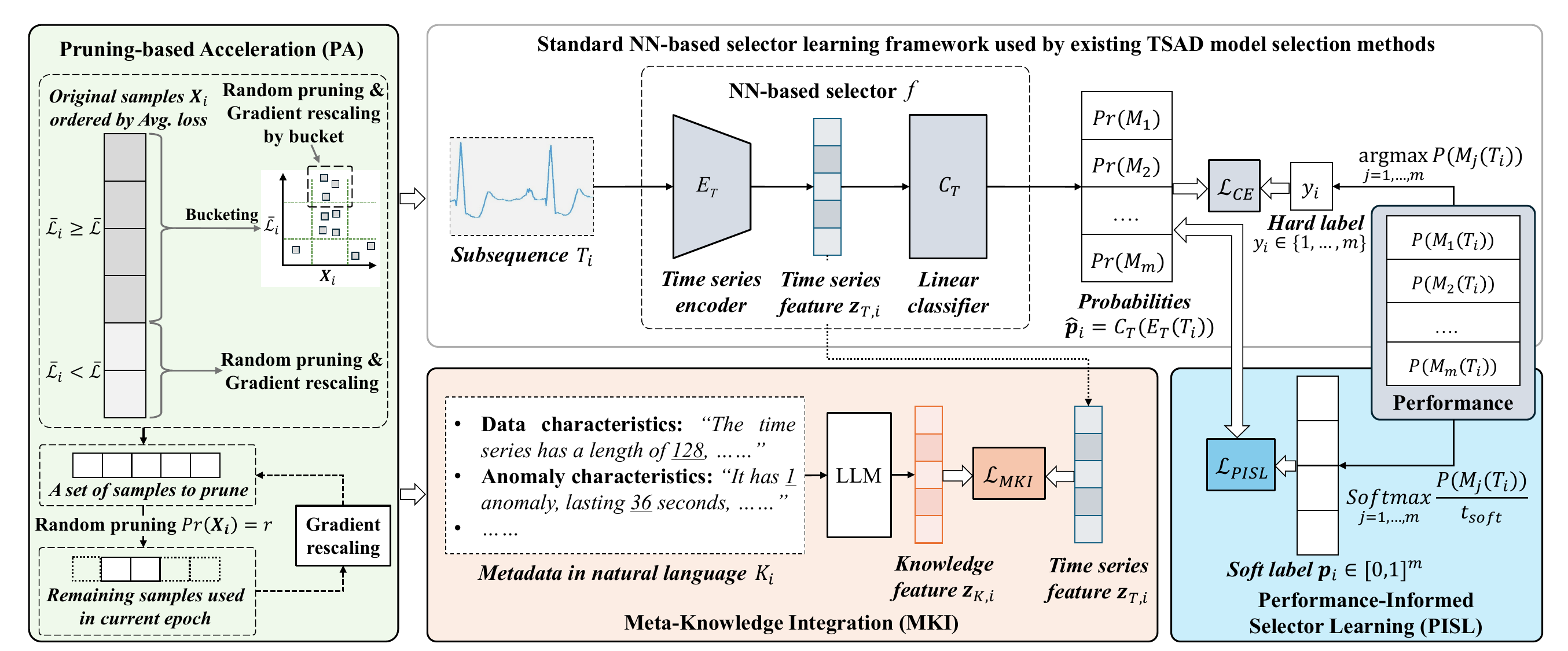}
    \vspace{-1.2ex}
    \caption{Overall framework of KDSelector}
    \label{fig:kdselector}
   % \vspace{-1.8ex}
\end{figure*}

The system provides a \textbf{Selector Management} module for users to easily save, manage, and load their learned selectors. Given a learned selector, the \textbf{Model Selection} module predicts the best model among the \textbf{TSAD Model Set} for each time series to detect. The selected model is run by the \textbf{Anomaly Detection} module and the detection results (e.g., anomaly score and overall performance) are visually shown to the users. We have now implemented 12 representative models in the TSAD model set following~\cite{ChooseWisely}. More models~\cite{schmidl2022anomaly} can be integrated in the same way in future work. 

\balance
\section{System Internals: \texttt{KDSelector}}\label{sec:internal}

Next, we introduce the proposed KDSelector, which is a general NN-based selector learning framework that aims to address the aforementioned challenges and serves as the internal of our system.  

\noindent
\textbf{\underline{Framework overview.}}
Fig.~\ref{fig:kdselector} illustrates the framework of KDSelector. Generally, \textit{an NN-based selector $f$ (i.e., TSC model) is a time series encoder $E_T$ appended by a linear classifier $C_T$}. For each input subsequence $T_i$, the selector first transforms it into a feature vector $\boldsymbol{z}_{T,i} = E_T(T_i)$. Then, the selector maps $\boldsymbol{z}_{T,i}$ to a vector $\hat{\boldsymbol{p}}_i=(Pr(M_1),\ldots,Pr(M_m)) = C_T(\boldsymbol{z}_{T,i})$ that represents the predicted probabilities of selecting the corresponding TSAD models, where the model with the highest probability is chosen.

As Fig.~\ref{fig:kdselector} shows, the \textit{\textbf{standard NN-based selector learning framework used by existing approaches}}~\cite{ChooseWisely} only uses the \textit{hard label $y_i = \arg\max_{j=1,\ldots,m}\ P(M_j(T_i))$ that represents the model with the best performance} to learn $f$, by minimizing the commonly used cross-entroy loss (denoted as $\mathcal{L}_{CE}$) between $\hat{\boldsymbol{p}}_i$ and $y_i$. Meanwhile, it requires iterating over all training samples at each epoch. As discussed in Sect.~\ref{sec:intro}, the above issues limit the performance of existing TSAD model selectors in terms of accuracy and training speed. To tackle these limitations, we design three plug-and-play modules that can be seamlessly integrated into the standard NN-based selector learning framework, which are described as follows.

\noindent
\textbf{\underline{Performance-informed selector learning (PISL).}} Considering that the detection performance $P(M_j(T_i)),\ j=1,\ldots,m$ not only indicates which model is the best for $T_i$ (i.e., $y_i$), but also reflects the complex relationship between the performance of all different models, we design PISL to make full use of the latter information to better train the selector. In principle, the TSAD model with better performance should have a higher probability of being selected. Thus, PISL \textit{transforms the performance scores into a probability distribution of selecting the corresponding models} using the \textit{Softmax} function, i.e., $\boldsymbol{p}_{i} = $\textit{Softmax}$_{j=1,\ldots,m}\ P(M_j(T_i))/t_{soft}$, where $t_{soft}$ is a hyperparameter that controls the smoothness of the distribution. The distribution $\boldsymbol{p}_{i}$ is used as a \textit{soft target (a.k.a. label)} to train $f$, which is achieved by \textit{minimizing the cross-entropy between the predicted distribution $\hat{\boldsymbol{p}}_i$ and the target $\boldsymbol{p}_{i}$}. Formally, the objective function is defined as $\mathcal{L}_{PISL} = \sum_{ i}\sum_{j=1}^m \boldsymbol{p}_{i,j}\log\hat{\boldsymbol{p}}_{i,j}$.

PISL can be integrated into existing NN-based selector learning methods regardless of NN architectures, by optimizing the objective $(1-\alpha)\mathcal{L}_{CE} + \alpha\mathcal{L}_{PISL}$, where $\alpha$ controls the relative importance of the soft label $\boldsymbol{p}_{i}$ and the hard label $y_i$.

\noindent
\textbf{\underline{Meta-knowledge integration (MKI).}} To gain knowledge from diverse metadata, MKI is designed to take natural languages (i.e., texts) as input to allow flexible and easy description of all kinds of metadata (e.g., the data and anomaly characteristics shown in Fig.~\ref{fig:kdselector}). The input, denoted as $K_i$, is then fed into a pre-trained LLM (e.g., BERT~\cite{bert}) to \textit{transform the text into a unified feature vector $\boldsymbol{z}_{K,i}$}, to take advantage of the superior ability of LLMs in natural language understanding. To integrate the knowledge in the metadata into the selector, from the perspective of information theory, we design a learning objective to \textit{maximize the mutual information (MI) between the features of the time series and the metadata}. It is achieved by mapping $\boldsymbol{z}_{T,i}$ and $\boldsymbol{z}_{K,i}$ into a shared space $\mathbb{R}^H$ using two projections $h_{T}$ and $h_{K}$, respectively, and then minimizing the InfoNCE loss~\cite{csl} (denoted as  $\mathcal{L}_{InfoNCE}$ ) that represents the opposite of a lower bound of MI between two random variables. The objective function is denoted as $\mathcal{L}_{MKI}=\mathcal{L}_{InfoNCE}\big(\{h_{T}(\boldsymbol{z}_{T,i}),h_{K}(\boldsymbol{z}_{K,i}) | \forall i\}\big)$. 

Similar to PISL, users can integrate MKI just by adding $\lambda\mathcal{L}_{MKI}$ to the total loss, where $\lambda$ is used to control the importance of MKI. 

\noindent
\textbf{\underline{Pruning-based acceleration (PA).}} To achieve data-efficient NN training, the state-of-the-art method, namely \textit{InfoBatch}~\cite{qin2024infobatch}, evaluates the importance of each sample $\boldsymbol{X}_i$ ($\boldsymbol{X}_i = \{T_i,\boldsymbol{z}_{K,i}\}$ with MKI and $T_i$ otherwise) using its average loss in the past epochs (denoted as $\bar{\mathcal{L}}_i$), and randomly prunes each less important sample (i.e., $\boldsymbol{X}_i$ if $\bar{\mathcal{L}}_i < \bar{\mathcal{L}}$ where $\bar{\mathcal{L}}$ is the average loss of all samples) with a probability $r$. It then iterates over the remaining samples in the current epoch, where for the samples of $\bar{\mathcal{L}}_i < \bar{\mathcal{L}}$, the gradients used for SGD are rescaled by multiplying $1/(1-r)$ to maintain that training on the pruned dataset is similar to training on the original one. 

\begin{table*}[htbp]
  \begin{minipage}[c]{0.3\linewidth}
          \centering
         \includegraphics[width=.99\linewidth]{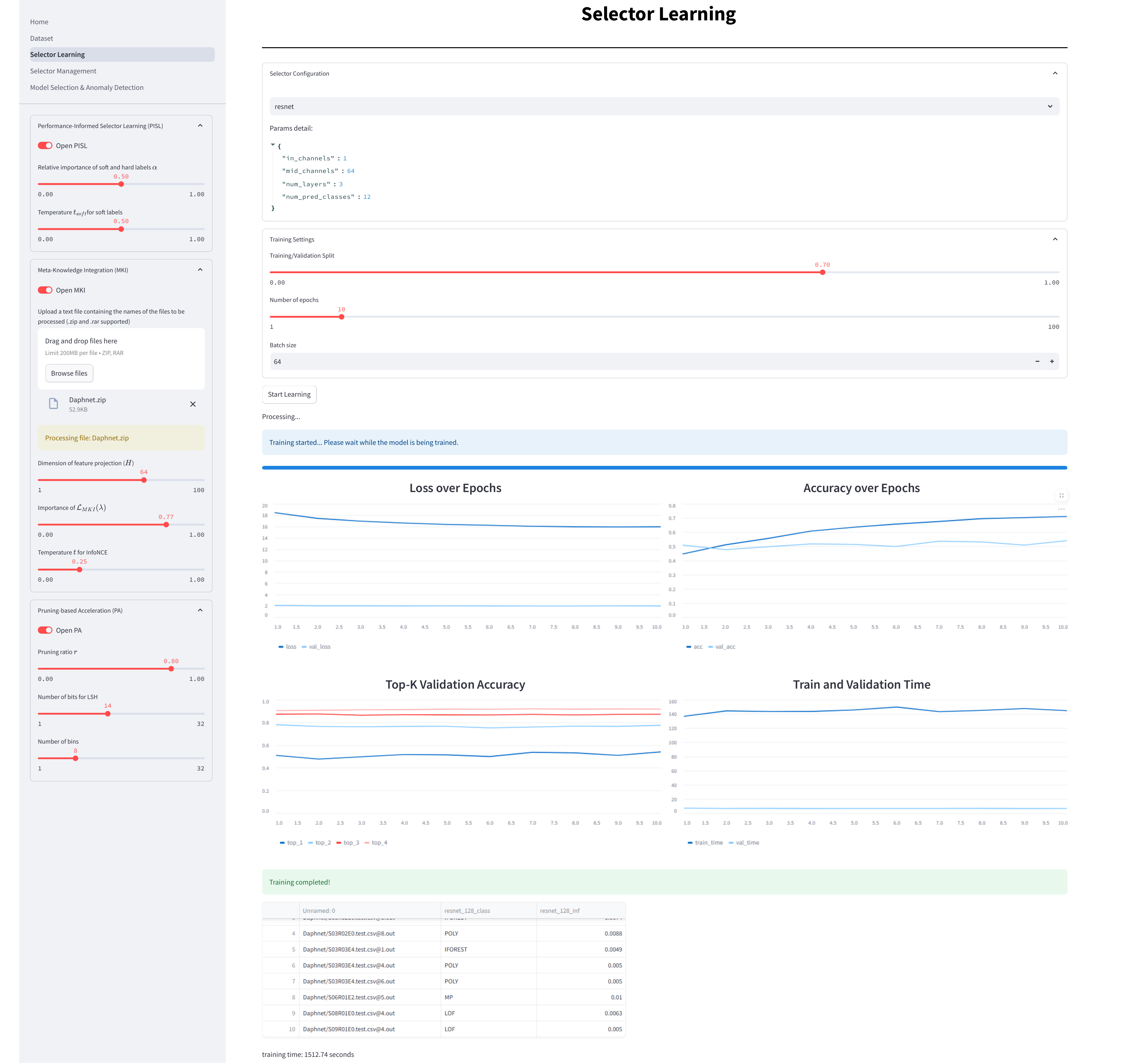}

\vspace{0.2ex}
\makeatletter\def\@captype{figure}\makeatother\caption{The system interfaces.}
         \label{fig:GUI}
 \end{minipage}%\hspace{1ex}
  \begin{minipage}[c]{0.31\linewidth}
          \centering
         \includegraphics[width=\linewidth]{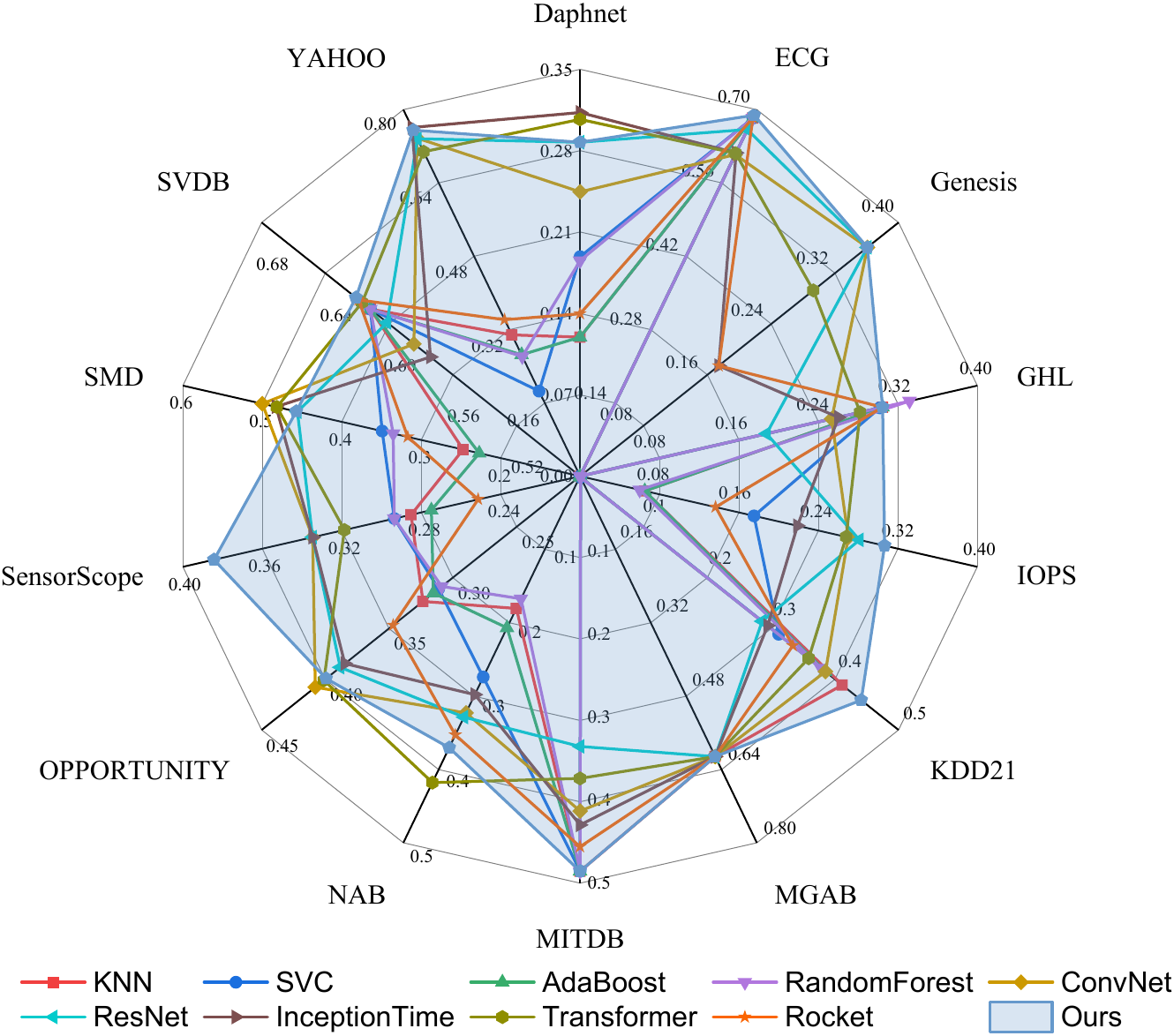}

\vspace{1ex}
\makeatletter\def\@captype{figure}\makeatother\caption{AUC-PR of different solutions.}
         \label{fig:overall_performance}
 \end{minipage}\hspace{1ex}
 \begin{minipage}[c]{0.3\linewidth}
 \centering
 \vspace{-4.3ex}
\makeatletter\def\@captype{table}\makeatother\caption{Results of PISL and MKI. (AUC-PR/total training time on all test/train sets. The same below.)}
\vspace{-3ex}
 \resizebox{.95\linewidth}{!}{
    \begin{tabular}{ccccc}
    \toprule
        Method & Standard & $+$ PISL & $+$ MKI & $+$ PISL \& MKI \\
    \midrule
        AUC-PR & 0.421 & 0.449 & 0.424 & \textbf{0.461}\\
        Time (mins) & 281.90 & \textbf{280.42} & 282.05 & 282.03\\
    \bottomrule
    \end{tabular}
    }
    \label{tab:knowledge}
\vspace{1ex}
    \makeatletter\def\@captype{table}\makeatother\caption{Results of PA on all datasets.}
    \vspace{-3ex}
     \resizebox{.9\linewidth}{!}{
    \begin{tabular}{cccc}
    \toprule
        Method & Full data & $+$ InfoBatch & $+$ PA (Ours) \\
    \midrule
        AUC-PR & \textbf{0.461} & 0.455$_{\downarrow0.006}$ & 0.452$_{\downarrow0.009}$\\
         Time (mins) & 282.03 & 171.73$_{\downarrow39.1\%}$ & \textbf{117.72}$_{\downarrow58.3\%}$\\
    \bottomrule
    \end{tabular}
    }
    \label{tab:pruning}
\vspace{1ex}
    \makeatletter\def\@captype{table}\makeatother\caption{Results of KDSelector on different architectures (all datasets).}
    \vspace{-3ex}
    \resizebox{.95\linewidth}{!}{
    \begin{tabular}{cccc}
    \toprule
        Architecture & ResNet & InceptionTime & Transformer \\
    \midrule
      Improved AUC-PR & 0.040 & 0.046 & 0.015\\
      Saved time (\%) & 58.3\% & 70.96\% & 74.17\% \\
    \bottomrule
    \end{tabular}
    }
    \label{tab:NN-archi}
    
 \end{minipage}
% \vspace{-4.7ex}
\end{table*}

 %\begin{table}[h]
 %   \centering
 %   \caption{Effectiveness of knowledge enhancement (PISL and MKI).}
%    \resizebox{.8\linewidth}{!}{
%    \begin{tabular}{ccccc}
%    \toprule
%        Method & Standard & $+$ PISL & $+$ MKI & $+$ PISL \& MKI \\
%    \midrule
%        AUC-PR & 0.421 & 0.449 & 0.424 & \textbf{0.461}\\
%        Time (mins) & 281.9 & \textbf{280.42} & 282.05 & 282.03\\
%    \bottomrule
%    \end{tabular}
%    }
%    \label{tab:knowledge}
%\end{table}

%\begin{table}[h]
%    \centering
%    \caption{Effectiveness of pruning-based acceleration (PA).}
%    \resizebox{.75\linewidth}{!}{
%    \begin{tabular}{cccc}
%    \toprule
%        Method & Full data & $+$ InfoBatch & $+$ PA (Ours) \\
%    \midrule
%        AUC-PR & \textbf{0.461} & 0.455$_{\downarrow0.006}$ & 0.452$_{\downarrow0.009}$\\
%         Time (mins) & 282.03 & 171.73$_{\downarrow39.1\%}$ & \textbf{117.72}$_{\downarrow58.3\%}$\\
%    \bottomrule
%    \end{tabular}
%    }
%    \label{tab:pruning}
%\end{table}

%\begin{table}[h]
%    \centering
%    \caption{Effectiveness on different architectures.}
%    \resizebox{.8\linewidth}{!}{
%    \begin{tabular}{cccc}
%    \toprule
%        Architecture & ResNet & InceptionTime & Transformer \\
%    \midrule
%      Improved AUC-PR & 0.040 & 0.045 & 0.013\\
%      Saved time (\%) & 58.3\% & 70.96\% & 74.17\% \\
%    \bottomrule
%    \end{tabular}
%    }
%    \label{tab:NN-archi}
%\end{table}

However, in the TSAD model selection problem, there can be redundant samples $\boldsymbol{X}_i$ for $\bar{\mathcal{L}}_i \ge \bar{\mathcal{L}}$ that cannot be pruned by InfoBatch. In specific, there may exist \textit{samples that are similar to each other and have similar average losses}. By our theoretical analysis (detailed in Sect. A.1), these samples have \textit{almost identical contributions for training the selector}. Thus, we propose \textbf{\textit{a novel strategy to prune these redundant samples to speed up the selector learning}}. The core idea is to \textit{divide the samples $\boldsymbol{X}_i$ of $\bar{\mathcal{L}}_i \ge \bar{\mathcal{L}}$ into different buckets, where the samples within each bucket are similar in both themselves and their average losses, and then perform pruning by bucket}. Considering that the values of the training samples are invariant during the training, we use local sensitive hashing~\cite{charikar2002similarity-LSH} (LSH) to \textit{efficiently hash all similar samples to the same hash tables before the training starts}. At each training epoch, we \textit{divide $\boldsymbol{X}_i$ of $\bar{\mathcal{L}}_i \ge \bar{\mathcal{L}}$ into $p$ equi-depth bins according to the current $\bar{\mathcal{L}}_i$, divide the samples that fall into the same hash table and bin into one bucket, and perform random pruning and gradient rescaling as InfoBatch for each bucket with more than one sample}. The samples of $\bar{\mathcal{L}}_i < \bar{\mathcal{L}}$ are pruned as InfoBacth without bucketing. We show that training using the proposed PA can still achieve a similar result to training without pruning (see Sect. A.2).   

Again, PA is general for accelerating the training of any NN-based selector. It is used as a plug-and-play module in our system.

\vspace{-0.5ex}
\balance
\section{Demonstration Scenarios}
In our demonstration, we intend to show how audiences achieve TSAD model selection using our system, and how the proposed KDSelector helps them improve the accuracy and training speed of the NN-based selectors.  We have prepared the 16 TSB-UAD~\cite{schmidl2022anomaly} benchmark datasets used in~\cite{ChooseWisely} for the audiences. They can also test on their own data using our system.

\noindent
\textbf{\underline{Pipeline for TSAD model selection.}} Fig.~\ref{fig:GUI} shows the interfaces of our system. In a nutshell, the user takes the following three steps to perform TSAD model selection by using the system.

\textbf{(1) Selector learning.} In this step, the user first uploads the historical data and configures the selector learning method. The system provides both NN-based and non-NN-based selectors. If the user chooses an NN-based selector, he/she can flexibly integrate the proposed modules in KDSelector into the learning. Once the user clicks on the "Start Learning" button, the system runs the learning method on the training data to learn the selector. The system provides visualization and evaluation functions for the user to validate the selector. It also allows the user to save and manage the learned selectors for easy reuse.

\textbf{(2) Model selection.}
At this stage, users upload their time series of interest and apply the selector learned above to predict the best TSAD model for each series. The system also shows the votes of different models to help users understand how the selection is made.

\textbf{(3) Anomaly detection.} After selection, the user can interact with the system to run the selected model on the corresponding time series and visually assess the detection results, such as the anomaly scores predicted by the TSAD model and the overall performance evaluated using a metric of interest.  The user can also run alternative models for a comparative analysis to validate the effectiveness of the model selection.

\noindent
\textbf{\underline{Superiority of KDSelector.}} Users can thoroughly evaluate our KDSelector. For example, by experimenting using the train/test data and settings following the benchmark~\cite{ChooseWisely} and reporting the accuracy (e.g., AUC-PR of the selected TSAD models) and training time, the user can draw three main conclusions: 
\textbf{(i) Effective knowledge-enhanced and data-efficient learning:} PISL and MKI can improve the accuracy of the learned selector with negligible training time overhead (see Table~\ref{tab:knowledge}), while PA can save more training time than using full data and InfoBatch with almost lossless accuracy (see Table~\ref{tab:pruning}).
\textbf{(ii) Architecture-agnostic:} KDSelector is effective for different selector architectures (see Table~\ref{tab:NN-archi}).
\textbf{(iii) Better model selection solution:} By integrating KDSelector into existing NN-based selectors, e.g., ResNet as evaluated in \texttt{Ours} in Fig.~\ref{fig:overall_performance}, the user can obtain a model selection solution (i.e., \texttt{Ours}) that outperforms existing solutions across different datasets and domains (see Fig.~\ref{fig:overall_performance}). We refer interested readers to Sect. B for the detailed setups and full results.

%\begin{acks}
%To Robert, for the bagels and explaining CMYK and color spaces.
%\end{acks}

%\vspace{-1.5ex}

\vspace{-0.5ex}
\normalem
\bibliographystyle{ACM-Reference-Format}
\bibliography{sample}

\appendix
\section{Theoretical Analysis}

This section shows the theoretical analysis results mentioned in Sect. 3 in detail, including the redundancy of the training samples that are similar to each other and have similar training losses in terms of their contributions to the selector learning, and the effectiveness of the proposed pruning-based acceleration in terms of achieving a similar result as training on full data. 

\subsection{Redundancy of Training Samples}
Denote $\mathcal{L}_i = \mathcal{L}(F(\boldsymbol{X_i};\boldsymbol{\Theta}))$ the loss of $\boldsymbol{X_i}$ at current epoch, where $F(\boldsymbol{X};\Theta)$ is the full model including both the selector $f$ and the projections $h_T$ and $h_K$. $\Theta$ is the set of parameters. Recall that NN-based selector is learned using SGD, by which the learning result (i.e., the update of the parameters) of each iteration depends on the gradient of $\mathcal{L}$ with respect to $\Theta$, i.e.,
\begin{equation}
    \Theta^{t+1} = \Theta^t - \eta \sum_i \nabla_\Theta \mathcal{L}_i, 
\end{equation}
where $t$ is the current epoch, $\eta$ is the learning rate, and $\nabla_\Theta \mathcal{L}_i$ is the gradient of $\mathcal{L}_i$ respect to $\Theta$. According to the chain rule, we have
\begin{equation}
    \nabla \mathcal{L}_i = \nabla_F \mathcal{L}_i \cdot \nabla_\Theta F_i,\label{eq:gradient_Li}
\end{equation}
where $F_i$ represents $F(\boldsymbol{X}_i)$.

Suppose $\boldsymbol{X}_i$ and $\boldsymbol{X}_j$ are two training samples that are similar in themselves and in their losses, i.e.,
\begin{equation}
||\boldsymbol{X}_i - \boldsymbol{X}_j|| < \delta_X,  \label{eq:diff_X}  
\end{equation}
and
\begin{equation}
    |\mathcal{L}_i - \mathcal{L}_j| < \delta_L,\label{eq:diff_L}
\end{equation}
where $\delta_X > 0$ and $\delta_L > 0$ are two small values. Based on Eq.~\eqref{eq:gradient_Li}, the difference of their contributions to the learning is
\begin{equation}
\begin{split}
        &||\nabla_\Theta \mathcal{L}_i - \nabla_\Theta \mathcal{L}_j|| \\=& ||\nabla_F \mathcal{L}_i \cdot \nabla_\Theta F_i - \nabla_F \mathcal{L}_j \cdot \nabla_\Theta F_j|| \\
        =& ||\nabla_F \mathcal{L}_i(\nabla_\Theta F_i - \nabla_\Theta F_j) + \nabla_\Theta F_j (\nabla_F \mathcal{L}_i - \nabla_F \mathcal{L}_j)||.
\end{split}
\end{equation}
By using the triangle inequality, we have
\begin{equation}
\begin{split}
    ||\nabla_\Theta \mathcal{L}_i - \nabla_\Theta \mathcal{L}_j|| \le& ||\nabla_F \mathcal{L}_i||\cdot||\nabla_\Theta F_i - \nabla_\Theta F_j|| \\ &+ ||\nabla_\Theta F_j||\cdot||\nabla_F \mathcal{L}_i - \nabla_F \mathcal{L}_j)||.\label{eq:inequality}
\end{split}
\end{equation}

By using SGD, we have to ensure that the gradient in Eq.~\eqref{eq:gradient_Li} is bounded (e.g., by gradient clipping), i.e., 
\begin{equation}
    ||\nabla_F \mathcal{L}_i|| \le B_L, \label{eq:bound_L}
\end{equation}
and
\begin{equation}
    ||\nabla_\Theta F_i|| \le B_F,\label{eq:bound_F}
\end{equation}
where $B_L$ and $B_F$ are the corresponding bounds. 

Assume that $\nabla_\Theta F$ and $\nabla_F \mathcal{L}$ are \textit{Lipschitz} for $\boldsymbol{X}$. According to Eq.~\ref{eq:diff_X} we have
\begin{equation}
    ||\nabla_\Theta F_i - \nabla_\Theta F_j|| \le C_F||\boldsymbol{X}_i - \boldsymbol{X}_i|| < C_F\delta_X,
\end{equation}
and
\begin{equation}
  ||\nabla_F \mathcal{L}_i - \nabla_F \mathcal{L}_j)||  \le C_L||\boldsymbol{X}_i - \boldsymbol{X}_i|| < C_L\delta_X,\label{eq:lipschitz_L}
\end{equation}
where $C_F > 0$ and $C_L > 0$ are two constants.

Substituting Eqs~\eqref{eq:bound_L}-\eqref{eq:lipschitz_L} back to Eq.~\eqref{eq:inequality}, we get
\begin{equation}
    ||\nabla_\Theta \mathcal{L}_i - \nabla_\Theta \mathcal{L}_j|| \le (B_LC_F + B_FC_L) ||\boldsymbol{X_i} - \boldsymbol{X_j}|| < (B_LC_F + B_FC_L)\delta_X.\label{eq:grad_bounded_by_X}
\end{equation}

From Eq.~\eqref{eq:grad_bounded_by_X}, we can see that $||\nabla_\Theta \mathcal{L}_i - \nabla_\Theta \mathcal{L}_j|| \to 0$ when $\delta_X \to 0$, indicating that the samples close to each other have a similar contribution for updating the parameters of $F$ (also including the selector $f$).

Moreover, assume that the loss is \textit{strongly convex} (e.g., by L2 regularization). We have
\begin{equation}
\begin{split}
  \nabla_F \mathcal{L}_j \nabla_\Theta F_j (\boldsymbol{X}_i - \boldsymbol{X}_j) + \frac{\mu}{2}||\boldsymbol{X}_i - \boldsymbol{X}_j||^2 \le  \mathcal{L}_i - \mathcal{L}_j ,\label{eq:strongly_convex}
\end{split}
\end{equation}
where $\mu > 0$ is a constant. Recall that $\nabla_F \mathcal{L}_j$ and $\nabla_\Theta F_j$ are bounded as Eqs.~\eqref{eq:bound_L}-\eqref{eq:bound_F}. Note that the second-order term with respect to $\boldsymbol{X}$ in Eq.~\eqref{eq:strongly_convex} can be ignored when $\delta_X \to 0$. Thus, based on Eq.\eqref{eq:diff_L}, when $|\nabla_F \mathcal{L}_j  \nabla_\Theta F_j (\boldsymbol{X}_i - \boldsymbol{X}_j)| \le |\mathcal{L}_i-\mathcal{L}_j|$ we get
\begin{equation}
    ||\boldsymbol{X}_i - \boldsymbol{X}_j|| \le A\delta_L,\label{eq:X_bounded_by_delta_L}
\end{equation}
where 
\begin{equation}
    A = \frac{1}{||\nabla_F \mathcal{L}_j\nabla_\Theta F_j||\cdot|\cos<\nabla_F \mathcal{L}_j\nabla_\Theta F_j,\boldsymbol{X}_i - \boldsymbol{X}_j>|}.
\end{equation}

By substituting Eq.~\eqref{eq:X_bounded_by_delta_L} to Eq.~\eqref{eq:grad_bounded_by_X}, we see that the similarity condition in the loss may provide a tighter bound for $||\nabla_\Theta \mathcal{L}_i - \nabla_\Theta \mathcal{L}_j||$ when $A\delta_L < \delta_X$. 

In conclusion, the samples that are similar both in themselves and in their losses have almost \textbf{\textit{identical contributions to the training}}. Note that in our PA module, we use the average loss over the last $t-1$ training epochs (i.e., $\bar{\mathcal{L}}_i$) to approximate the loss in the current epoch (i.e., $\mathcal{L}_i$) following~\cite{qin2024infobatch} to achieve efficient pruning. 
%by applying the first-order Taylor approximation to the term $\nabla_F L_i - \nabla_F L_j$ in Eq.~\ref{eq:inequality}, we get
%\begin{equation}
%    \nabla_F L_i - \nabla_F L_j \approx H_F(\boldsymbol{X}_j)(F_i - F_j),
%\end{equation}
%where $H_F$ is the Hessian matrix of $\mathcal{L}$ with respect to $F$. 
\subsection{Effectiveness of Pruning-based Acceleration}
Recall that in our proposed PA, we perform random pruning and gradient rescale for the samples of $\bar{\mathcal{L}}_i \ge \bar{\mathcal{L}}$ and falling in the same buckets, and the samples of $\bar{\mathcal{L}}_i < \bar{\mathcal{L}}$. Therefore, we can divide the full dataset, denoted as $\mathcal{D}$, into $2$ disjoint subsets, including $\mathcal{D}_{1}$ that combines all buckets that need to prune and $\mathcal{D}_{2}$ other. Formally, we have
\begin{equation}
    \mathcal{D} = \mathcal{D}_{1}\cup \mathcal{D}_{2}.
\end{equation}

Denote $S_1$ the subset of $\mathcal{D}_{1}$ after pruning and $\mathcal{S}$ the pruned dataset used for the current training epoch. We have
\begin{equation}
    \mathcal{S} = \mathcal{S}_1 \cup \mathcal{D}_{2}.
\end{equation}

For each sample $\boldsymbol{X}_i$ to prune, the probability of pruning it is $r$, which is formulated as
\begin{equation}
    Pr(\boldsymbol{X}_i) = r.
\end{equation}

With gradient rescaling, the losses for the remaining samples in the pruned subsets, i.e., $\mathcal{S}_1$, are multiplied by the factor $1/(1-r)$, which is the same as multiplying the corresponding gradients by the factor. Assume that all samples $\boldsymbol{X}$ are drawn from a continuous distribution $\rho(\boldsymbol{X})$. The objective of selector learning on the full dataset can be formulated as
\begin{equation}
    \underset{\Theta}{\arg\min} \underset{\boldsymbol{X} \in \mathcal{D}}{\mathbb{E}}[\mathcal{L}(\boldsymbol{X};\Theta)] =  \int_{\boldsymbol{X}\in \mathcal{D}} \mathcal{L}(\boldsymbol{X;\Theta})\rho(\boldsymbol{X})d\boldsymbol{X}.\label{eq:objective_full_data}
\end{equation}

After pruning using the proposed PA, the selector learning objective becomes
\begin{equation}
\begin{split}
 \ \ \ \ \ \ \ \ \ \ \ &\underset{\Theta}{\arg\min} \underset{\boldsymbol{X} \in \mathcal{S}}{\mathbb{E}}[\mathcal{L}^\prime(\boldsymbol{X};\Theta)]  \\ =& \int_{\boldsymbol{X}\in \mathcal{S}_1} \frac{1}{1-r}\mathcal{L}(\boldsymbol{X;\Theta})\rho(\boldsymbol{X})d\boldsymbol{X} +  \int_{\boldsymbol{X}\in \mathcal{D}_2} \mathcal{L}(\boldsymbol{X;\Theta})\rho(\boldsymbol{X})d\boldsymbol{X},  \label{eq:objective_pruned} 
\end{split}
\end{equation}
where the first term on the right-hand side can be derived to
\begin{equation}
\begin{split}
    &\ \ \ \ \int_{\boldsymbol{X}\in \mathcal{S}_1} \frac{1}{1-r}\mathcal{L}(\boldsymbol{X;\Theta})\rho(\boldsymbol{X})d\boldsymbol{X} \\
    &= \int_{\boldsymbol{X}\in \mathcal{D}_1} \frac{1-Pr(\boldsymbol{X})}{1-r}\mathcal{L}(\boldsymbol{X;\Theta})\rho(\boldsymbol{X})d\boldsymbol{X}\\
    &= \int_{\boldsymbol{X}\in \mathcal{D}_1} \mathcal{L}(\boldsymbol{X;\Theta})\rho(\boldsymbol{X})d\boldsymbol{X}.\label{eq:objective_pruned_subset}
\end{split}
\end{equation}
By substituting Eq.~\eqref{eq:objective_pruned_subset} to Eq.~\eqref{eq:objective_pruned}, we get
\begin{equation}
\begin{split}
  \ \ \ \ \ \ \ \ \ \ \ &\underset{\Theta}{\arg\min} \underset{\boldsymbol{X} \in \mathcal{S}}{\mathbb{E}}[\mathcal{L}^\prime(\boldsymbol{X};\Theta)] \\
    =& \int_{\boldsymbol{X}\in \mathcal{D}_1} \mathcal{L}(\boldsymbol{X;\Theta})\rho(\boldsymbol{X})d\boldsymbol{X} + \int_{\boldsymbol{X}\in \mathcal{D}_2} \mathcal{L}(\boldsymbol{X;\Theta})\rho(\boldsymbol{X})d\boldsymbol{X} \\
    =& \int_{\boldsymbol{X}\in \mathcal{D}} \mathcal{L}(\boldsymbol{X;\Theta})\rho(\boldsymbol{X})d\boldsymbol{X},
\end{split}
\end{equation}
which is consistent with the objective of learning on the full data without pruning as shown in Eq.~\eqref{eq:objective_full_data}. Therefore, we conclude that \textbf{\textit{learning using the proposed pruning strategy can achieve a similar result as learning on the full data}}.

\begin{table*}[h]
    \centering
            \caption{Dataset description~\cite{schmidl2022anomaly}.}
             \resizebox{.9\linewidth}{!}{
    \begin{tabular}{l|c}
     \toprule
      \textbf{Dataset}  &  \textbf{Description (Domain knowledge)}\\
     \midrule
Dodgers	& \makecell{is a loop sensor data for the Glendale on-ramp for the 101 North freeway in Los Angeles \\and the anomalies represent unusual traffic after a Dodgers game.}\\
\hline
ECG	&\makecell{is a standard electrocardiogram dataset and the anomalies represent ventricular premature \\contractions. We split one long series (MBA\_ECG14046) with length about 1e7 to 47 series\\ by first identifying the periodicity of the signal.}\\
\hline
IOPS&	\makecell{is a dataset with performance indicators that reflect the scale, quality of web services, and \\health status of a machine.}\\
\hline
KDD21&	\makecell{is a composite dataset released in a recent SIGKDD 2021 competition with 250 time series.}\\
\hline
MGAB&	\makecell{is composed of Mackey-Glass time series with non-trivial anomalies. Mackey-Glass time \\series exhibit chaotic behavior that is difficult for the human eye to distinguish.}\\
\hline
NAB&	\makecell{is composed of labeled real-world and artificial time series including AWS server metrics, \\online advertisement clicking rates, real time traffic data, and a collection of Twitter \\mentions of large publicly-traded companies.}\\
\hline
SensorScope&	\makecell{is a collection of environmental data, such as temperature, humidity, and solar radiation, \\collected from a typical tiered sensor measurement system.}\\
\hline
YAHOO&	\makecell{is a dataset published by Yahoo labs consisting of real and synthetic time series based on \\the real production traffic to some of the Yahoo production systems.}\\
\hline
Daphnet&	\makecell{contains the annotated readings of 3 acceleration sensors at the hip and leg of Parkinson’s \\disease patients that experience freezing of gait (FoG) during walking tasks.}\\
\hline
GHL&	\makecell{is a Gasoil Heating Loop Dataset and contains the status of 3 reservoirs such as the \\temperature and level. Anomalies indicate changes in max temperature or pump frequency.}\\
\hline
Genesis&	\makecell{is a portable pick-and-place demonstrator which uses an air tank to supply all the gripping \\and storage units.}\\
\hline
MITDB&	\makecell{contains 48 half-hour excerpts of two-channel ambulatory ECG recordings, obtained \\from 47 subjects studied by the BIH Arrhythmia Laboratory between 1975 and 1979.}\\
\hline
OPPORTUNITY&	\makecell{is a dataset devised to benchmark human activity recognition algorithms (e.g., \\classiffication, automatic data segmentation, sensor fusion, and feature extraction). \\The dataset comprises the readings of motion sensors recorded while users executed \\typical daily activities.}\\
\hline
Occupancy&	\makecell{contains experimental data used for binary classiffication (room occupancy) from \\temperature, humidity, light, and CO2. Ground-truth occupancy was obtained from \\time stamped pictures that were taken every minute.}\\
\hline
SMD&	\makecell{is a 5-week-long dataset collected from a large Internet company. This dataset \\contains 3 groups of entities from 28 different machines.}\\
\hline
SVDB&	\makecell{includes 78 half-hour ECG recordings chosen to supplement the examples of \\supraventricular arrhythmias in the MIT-BIH Arrhythmia Database.}\\
    \bottomrule
    \end{tabular}
    }
    \label{tab:datasets}
\end{table*}

\begin{table*}[]
    \centering
        \caption{TSAD models used for model selection~\cite{ChooseWisely,schmidl2022anomaly}.}
    \begin{tabular}{l|c}
    \toprule
    \textbf{TSAD model} & \textbf{Description}\\
    \midrule
  Isolation Forest (IForest) & \makecell{This method constructs the binary tree based on the space splitting and the \\nodes with shorter path lengths to the root are more likely to be anomalies.}\\
  \hline
  IForest1 & same as IForest, but each data point (individually) are used as input.\\
  \hline
The Local Outlier Factor (LOF)	& \makecell{This method computes the ratio of the neighboring density to the local density.}\\
\hline
The Histogram-based Outlier Score (HBOS) &	\makecell{This method constructs a histogram for the data and the inverse of the height\\ of the bin is used as the outlier score of the data point.}\\
\hline
Matrix Profile (MP) &	\makecell{This method calculates as anomaly the subsequence with the most significant \\1-NN distance.}\\
\hline
NORMA &	\makecell{This method identifies the normal pattern based on clustering and calculates \\each point's effective distance to the normal pattern.}\\
\hline
Principal Component Analysis (PCA)	& \makecell{This method projects data to a lower-dimensional hyperplane, and data points\\ with a significant distance from this plane can be identified as outliers.}\\
\hline
Autoencoder (AE) &	\makecell{This method projects data to the lower-dimensional latent space and reconstructs\\ the data, and outliers are expected to have more evident reconstruction deviation.}\\
\hline
LSTM-AD&	\makecell{This method build a non-linear relationship between current and previous time\\ series (using Long-Short-Term-Memory cells), and the outliers are detected \\by the deviation between the predicted and actual values.}\\
\hline
Polynomial Approximation (POLY)&	\makecell{This method build a non-linear relationship between current and previous \\time series (using polynomial decomposition), and the outliers are detected \\by the deviation between the predicted and actual values.}\\
\hline
CNN	& \makecell{This method build a non-linear relationship between current and previous \\time series (using convolutional Neural Network), and the outliers are \\detected by the deviation between the predicted and actual values.}\\
\hline
One-class Support Vector Machines (OCSVM) &	\makecell{This method fits the dataset to find the normal data's boundary.}\\
\bottomrule
    \end{tabular}

    \label{tab:TSAD_models}
\end{table*}

\section{Experiments}
This section shows the experimental setups and results in detail.

\subsection{Experimental Setups}

\vspace{0.5ex}
\noindent
\underline{\textbf{Datasets.}} We use the 16 TSB-UAD~\cite{schmidl2022anomaly} subsets following~\cite{ChooseWisely}, as described in Table~\ref{tab:datasets}. For a fair comparison, we use the recommended train/test split~\cite{ChooseWisely}, where the training set is a combination of samples from all 16 datasets, while the time series from 14 subsets are used for test as shown in Fig.~\ref{fig:overall_performance}.

\vspace{0.5ex}
\noindent
\underline{\textbf{Baselines.}} The baseline TSAD model selection solutions used for comparison, as shown in Fig.~\ref{fig:overall_performance}, are representative approaches chosen from~\cite{ChooseWisely} that have shown competitive model selection performance. These solutions can be divided into two categories.
\begin{itemize}
    \item \textbf{Non-NN-based methods.} This includes (i) \textit{feature-based methods} that use the open-source tool TSFresh to extract features from the input time series, and train traditional machine learning classifiers on top of the features. The classifiers contain K nearest neighbors (\texttt{KNN}), support vector classifier (\texttt{SVC}), \texttt{AdaBoost}, and random forest classifier (\texttt{RandomForest}), and (ii) \textit{kernel-based method} that refers to MiniRocket (abbreviated as \texttt{Rocket}) that uses multiple convolutional kernels generated at random in conjunction with a Ridge regression classifier.   
       \item \textbf{NN-based methods.} This includes three convolution-based models, i.e., \texttt{ConvNet} that uses convolutional layers to learn spatial features, \texttt{ResNet} that uses ConvNet with residual connections, and \texttt{InceptionTime} that combines ResNets with kernels of multiple sizes, and an advanced \texttt{Transformer} architecture that corresponds to SiT-stem in~\cite{ChooseWisely}.
\end{itemize}
We directly use the implementations open-sourced by~\cite{ChooseWisely} with their default parameters without specification for the baseline solutions. To achieve strong baselines, we run each baseline method using different subsequence lengths as $L \in \{16,32,64,128,256,512,768,1024\}$ and report the best result on each dataset. All the evaluated baseline methods have been implemented in the current system.

\vspace{0.5ex}
\noindent
\underline{\textbf{TSAD models}}. To achieve a fair comparison, we use the 12 representative TSAD models chosen by~\cite{ChooseWisely} as the candidates in our TSAD model set. The models are described in Table~\ref{tab:TSAD_models}. We use the open-source implementations and the default settings following~\cite{ChooseWisely}. All the TSAD methods have been integrated into our system.

\vspace{0.5ex}
\noindent
\underline{\textbf{Implementation details of KDSelector}}. 
The proposed KDSelector is implemented using Python 3.8 and PyTorch 1.12. The experiment was run on a server with Platinum 8260 CPUs and Ubuntu 20.04 LTS, using a single NVIDIA GTX 3090 GPU. The default selector architecture is ResNet during our experiments, while InceptionTime and Transformer are also used to validate the effectiveness on different architectures. We keep the settings of KDSelector consistent with its underlying selector architecture for a fair comparison. 

We use the base version of BERT~\cite{bert} for text embedding, of which the parameters are frozen during the selector learning. The metadata used for MKI includes the length of the input series, the number of anomalies the series contains, the lasting time of these anomalies, and the description of application domain of the dataset (see Table~\ref{tab:datasets}). The following \textbf{\textit{template is used to describe the metadata.}}

\vspace{1ex}
``\textit{This is a time series from dataset [Dataset name], [Description as Table~\ref{tab:datasets}]. The length of the series is [Length of series]. There are [Number of anomalies] anomalies in this series. The lengths of the anomalies are [Length of anomalies] (without this sentence if the number of anomalies is 0).}''
\vspace{1ex}

The projections $h_T$ and $h_K$ used in MKI are implemented using two multi-layer perceptions (MLPs), respectively. Each MLP has one hidden layer of 256 dimensions, with ReLU as the activation function. The output dimension $H$ is selected from $\{64, 256\}$. 

The hyper-parameters of PISL and MKI, including $t_{soft}$, $\alpha$, and $\lambda$ are selected from $\{0.2, 0.22, 0.25\}$, $\{0.2,0.4,1.0\}$, and $\{0.78, 1.0\}$, respectively. The temperature for the InfoNCE loss is set to 0.1. 

For PA evaluation, we set the pruning ratio $r$ to 0.8 (the same for InfoBatch). The number of bits used in LSH is set to 14, and the number of bins $p$ is set to 8. Other settings are the same as InfoBatch.    

\vspace{0.5ex}
\noindent
\underline{\textbf{Metrics.}} As shown in Sect. 4, we use AUC-PR to measure the accuracy of the selector. It is obtained by running the selected TSAD model on the corresponding time series and computing the metric using the true anomalies and the predicted anomaly scores of each data point. For training speed evaluation, we report the running time of the selector learning algorithm on the training dataset.

For a fair comparison, we exclude PA when comparing with existing solutions because they do not use any pruning strategy by default (Table~\ref{tab:knowledge}, AUR-PR in Table~\ref{tab:NN-archi}, and Fig.~\ref{fig:overall_performance}), while to evaluate PA, we keep the proposed PISL and MKI in use to compare the learning results using different pruning strategies (Table~\ref{tab:pruning} and saved time in Table~\ref{tab:NN-archi}).

\subsection{Full Results}
The full experimental results corresponding to Tables~\ref{tab:knowledge}-\ref{tab:NN-archi} and Fig.~\ref{fig:overall_performance} are shown as follows.

\begin{table*}[h]
    \centering
\vspace{6ex}
       \caption{Full results of PISL and MKI.}
    \begin{tabular}{lcccc}
        \toprule
        \textbf{Method} & Standard & $+$ PISL & $+$ MKI & $+$ PISL \& MKI \\
        \midrule
Daphnet	&0.2873	&0.2873	&\textbf{0.3014}	&0.2873\\
ECG	&0.6624	&\textbf{0.6897}	&0.6259	&\textbf{0.6897}\\
Genesis	&\textbf{0.3617}	&\textbf{0.3617}	&\textbf{0.3617}	&\textbf{0.3617}\\
GHL	&0.1932	&\textbf{0.3071}	&0.2303	&0.3035\\
IOPS	&0.2843	&0.2843	&\textbf{0.309}	&\textbf{0.309}\\
KDD21	&0.2902	&0.3875	&0.3125	&\textbf{0.4426}\\
MGAB	&\textbf{0.614}	&\textbf{0.614}	&\textbf{0.614}	&\textbf{0.614}\\
MITDB	&0.3355	&\textbf{0.4856}	&0.3123	&\textbf{0.4856}\\
NAB	&\textbf{0.3319}	&\textbf{0.3319}	&0.3137	&0.3279\\
OPPORTUNITY	&0.3886	&0.3886	&\textbf{0.4031}	&0.3995\\
SensorScope	&0.335	&0.335	&0.335	&\textbf{0.3844}\\
SMD	&0.4561	&0.4561	&0.4501	&\textbf{0.4576}\\
SVDB	&0.6212	&0.6212	&0.6215	&\textbf{0.6337}\\
YAHOO	&0.737	&0.737	&0.7535	&\textbf{0.7558}\\

        \hline
        Average AUC-PR  & 0.421 & 0.449 & 0.424 & \textbf{0.461}\\
        \midrule
        Total training Time (mins) & 281.9 & \textbf{280.42} & 282.05 & 282.03\\
        \bottomrule
    \end{tabular}
    \label{tab:knowledge_full}
\end{table*}

\begin{table*}[h]
    \centering
       \caption{Full results of PA.}
    \begin{tabular}{lccc}
        \toprule
        \textbf{Method} & Full data & $+$ InfoBatch & $+$ PA (Ours) \\
        \midrule
Daphnet	&0.2873	&0.2724	&\textbf{0.2933}\\
ECG	&\textbf{0.6897}	&\textbf{0.6897}	&\textbf{0.6897}\\
Genesis	&\textbf{0.3617}	&\textbf{0.3617}	&\textbf{0.3617}\\
GHL	&0.3035	&\textbf{0.3071}	&\textbf{0.3071}\\
IOPS	&0.3090	&0.2843	&\textbf{0.3762}\\
KDD21	&\textbf{0.4426}	&0.4304	&0.3941\\
MGAB	&\textbf{0.6140}	&\textbf{0.6140}	&\textbf{0.6140}\\
MITDB	&\textbf{0.4856}	&\textbf{0.4856}	&\textbf{0.4856}\\
NAB	&0.3279	&0.3398	&\textbf{0.3643}\\
OPPORTUNITY	&\textbf{0.3995}	&0.3725	&0.3928\\
SensorScope	&\textbf{0.3844}	&\textbf{0.3844}	&0.3132\\
SMD	&0.4576	&\textbf{0.4602}	&0.4277\\
SVDB	&\textbf{0.6337}	&0.6213	&0.6118\\
YAHOO	&\textbf{0.7558}	&0.7481	&0.7070\\
        \hline
      Average AUC-PR & \textbf{0.461} & 0.455$_{\downarrow0.006}$ & 0.452$_{\downarrow0.009}$\\
      \midrule
         Total training time (mins) & 282.03 & 171.73$_{\downarrow39.1\%}$ & \textbf{117.72}$_{\downarrow58.3\%}$\\
        \bottomrule
    \end{tabular}
    \label{tab:PA_full}
\end{table*}

\begin{table*}[h]
    \centering
    \vspace{6ex}
       \caption{Full results on different architectures.}
    \begin{tabular}{l|cc|cc|cc}
        \toprule
        \textbf{Architecture} & \multicolumn{2}{c|}{ResNet} & \multicolumn{2}{c|}{InceptionTime} & \multicolumn{2}{c}{Transformer} \\
        \hline
        
       \rule{0pt}{3ex}\textbf{Method} & Default & $+$ KDSelector & Default & $+$ KDSelector & Default & $+$ KDSelector \\
        \midrule
Daphnet	      &\textbf{0.2873}	&\textbf{0.2873}	  &\textbf{0.3129}	&0.2990	                          &\textbf{0.3070}	             &0.2804\\
ECG	          &0.6624	        &\textbf{0.6897}	  &0.6187	        &\textbf{0.6917}	              &0.6176	                     &\textbf{0.6897}\\
Genesis	      &\textbf{0.3617}	&\textbf{0.3617}	  &0.1796	        &\textbf{0.3617}	              &0.2957	                     &\textbf{0.3617}\\
GHL	          &0.1932	        &\textbf{0.3035}	  &0.2637	        &\textbf{0.3071}	              &0.2851	                     &\textbf{0.3349}\\
IOPS	      &0.2843	        &\textbf{0.3090}	  &0.2235	        &\textbf{0.3045}	              &0.2714	                     &\textbf{0.2845}\\
KDD21	      &0.2902	        &\textbf{0.4426}	  &0.2987	        &\textbf{0.4010}	              &0.3616	                     &\textbf{0.3799}\\
MGAB	      &\textbf{0.6140}	&\textbf{0.6140}	  &\textbf{0.6140}	&\textbf{0.6140}	              &0.6140	                     &\textbf{0.6140}\\
MITDB	      &0.3355	        &\textbf{0.4856}	  &0.4298	        &\textbf{0.5089}	              &0.3739	                     &\textbf{0.4856}\\
NAB	          &\textbf{0.3319}	&0.3279	              &0.3019	        &\textbf{0.3701}	              &\textbf{0.4195}	             &0.3530\\
OPPORTUNITY	  &0.3886	        &\textbf{0.3995}	  &\textbf{0.3849}	&0.3842	                          &\textbf{0.4013}	             &0.3656\\
SensorScope   &0.3350	        &\textbf{0.3844}	  &0.3350	        &\textbf{0.3545}	              &0.3191	                     &\textbf{0.3312}\\
SMD	          &0.4561	        &\textbf{0.4576}	  &\textbf{0.4832}	&0.4684	                          &\textbf{0.4819}	             &0.4444\\
SVDB	      &0.6212	        &\textbf{0.6337}	  &0.5940	        &\textbf{0.6391}	              &0.6369	                     &\textbf{0.6372}\\
YAHOO	      &0.7370	        &\textbf{0.7558}	  &\textbf{0.7616}	&0.7457	                          &0.7091	                     &\textbf{0.7384}\\

        \hline
    Average AUC-PR & 0.4213	    &\textbf{0.4609}	  &0.4144	&\textbf{0.4607}	&0.4353	&\textbf{0.4500}\\
 \cline{2-7}
 Improved AUC-PR & \multicolumn{2}{c|}{0.0396} & \multicolumn{2}{c|}{0.0463} & \multicolumn{2}{c}{0.0147}\\
\hline
 Total training time (mins) & 282.03 & \textbf{117.72} & 292.99 & \textbf{85.09} & 343.94 & \textbf{88.85}\\
 \cline{2-7}
Saved time (\%) & \multicolumn{2}{c|}{58.3\%} & \multicolumn{2}{c|}{70.96\%} & \multicolumn{2}{c}{74.17\%}\\
        \bottomrule
    \end{tabular}
    \label{tab:architecture_full}
\end{table*}

\begin{table*}[]
    \centering
        \caption{Full results of different model selection solutions.}
    \begin{tabular}{lcccccccccc}
    \toprule
\textbf{Method}	&KNN	&SVC	&AdaBoost	&RandomForest	&ConvNet	&ResNet	&InceptionTime	&Transformer	&Rocket	&\textbf{Ours}\\
\midrule
Daphnet	    &0.1197	  &0.1888	&0.1197	  &0.1854	&0.2445	  &0.2873	&\textbf{0.3129}	  &0.3070	&0.1397	  &0.2873\\
ECG	        &0.6842	  &0.6842	&0.6842	  &0.6842	&0.6154	  &0.6624	&0.6187	  &0.6176	&0.6842	  &\textbf{0.6897}\\
Genesis	    &0.0017	  &0.0017	&0.0017	  &0.0017	&\textbf{0.3617}	  &\textbf{0.3617}	&0.1796	  &0.2957	&0.1796	  &\textbf{0.3617}\\
GHL	        &0.3071	  &0.3042	&0.3071	  &\textbf{0.3335}	&0.2571	  &0.1932	&0.2637	  &0.2851	&0.3068	  &0.3071\\
IOPS	    &0.0734	  &0.1807	&0.0734	  &0.0686	&0.2732	  &0.2843	&0.2235	  &0.2714	&0.1430	  &\textbf{0.3090}\\
KDD21	    &0.4132	  &0.3154	&0.3876	  &0.3808	&0.3874	  &0.2902	&0.2987	  &0.3616	&0.3372	  &\textbf{0.4426}\\
MGAB	    &\textbf{0.6140}	  &0.0122	&0.0122	  &0.0122	&\textbf{0.6140}	  &\textbf{0.6140}	&\textbf{0.6140}	  &\textbf{0.6140}	&\textbf{0.6140}	  &\textbf{0.6140}\\
MITDB	    &\textbf{0.4856}	  &\textbf{0.4856}	&\textbf{0.4856}	  &\textbf{0.4856}	&0.4132	  &0.3355	&0.4298	  &0.3739	&0.4562	  &\textbf{0.4856}\\
NAB	        &0.1867	  &0.2783	&0.2127	  &0.1739	&0.3263	  &0.3319	&0.3019	  &\textbf{0.4195}	&0.3558	  &0.3730\\
OPPORTUNITY	&0.3238	  &0.3116	&0.3157	  &0.3092	&\textbf{0.4080}	  &0.3886	&0.3849	  &0.4013	&0.3472	  &0.3995\\
SensorScope	&0.2857	  &0.2941	&0.2755	  &0.2941	&0.3350	  &0.3350	&0.3350	  &0.3191	&0.2521	  &\textbf{0.3844}\\
SMD	        &0.2479	  &0.3497	&0.2278	  &0.3358	&\textbf{0.5004}	  &0.4561	&0.4832	  &0.4819	&0.3175	  &0.4576\\
SVDB	    &0.6317	  &0.6317	&0.6317	  &0.6317	&0.6043	  &0.6212	&0.5940	  &0.6369	&0.6391	  &\textbf{0.6408}\\
YAHOO	    &0.3144	  &0.1934	&0.2718	  &0.2681	&0.7385	  &0.7370	&\textbf{0.7616}	  &0.7091	&0.3474	  &0.7558\\
\bottomrule
    \end{tabular}

    \label{tab:overall_performance_all}
\end{table*}

\end{document}